\documentclass[fullpaper]{nldl}

\renewcommand{\DOI}{12345}
\usepackage{amsmath}
\usepackage[utf8]{inputenc}
\usepackage{url}
\usepackage{graphicx}
\usepackage{subcaption}
\usepackage{authblk}

\usepackage{algorithm}
\usepackage{accents}
\usepackage{algorithmic}

\usepackage{amsfonts}
\usepackage[square,numbers]{natbib}

\usepackage{breakurl}
\usepackage[breaklinks]{hyperref}

\usepackage{hyperref}

\title{Approaching adverse event detection utilizing transformers on clinical time-series}
\author[1]{Helge Fredriksen\thanks{Corresponding Author: helge.fredriksen@uit.no}}
\author[2]{Per Joel Burman}
\author[2]{Ashenafi Woldaregay}
\author[2]{Karl Øyvind Mikalsen}
\author[3]{Ståle Nymo}
\affil[1]{UiT - The Arctic University of Norway}
\affil[2]{The Norwegian Centre for Clinical Artificial Intelligence}
\affil[3]{Nordland Hospital Trust}

\date{\vspace{-5ex}}

\begin{document}
\nldlmaketitle

\begin{abstract}  
Patients being admitted to a hospital will most often be associated with a certain clinical development during their stay. However, there is always a risk of patients being subject to the wrong diagnosis or to a certain treatment not pertaining to the desired effect, potentially leading to adverse events. Our research aims to develop an anomaly detection system for identifying deviations from expected clinical trajectories.
To address this goal we analyzed 16 months of vital sign recordings obtained from the Nordland Hospital Trust (NHT). We employed an self-supervised framework based on the STraTS transformer architecture to represent the time series data in a latent space. These representations were then subjected to various clustering techniques to explore potential patient phenotypes based on their clinical progress. While our preliminary results from this ongoing research are promising, they underscore the importance of enhancing the dataset with additional demographic information from patients. This additional data will be crucial for a more comprehensive evaluation of the method's performance.

\end{abstract}

\section{Background}
As the precision, efficiency and availability of treatments keeps improving, prioritizing resources to maximize the benefit for patients becomes a vital element of a functioning health care system. An important aspect of such prioritization is the identification of patients at increased risk of adverse events, in order to provide appropriate evaluation and care to these patients.
For in-patient monitoring, the National Early Warning Score (NEWS) has recently been recommended for implementation in all Norwegian hospitals  \cite{HelseDirektoratet} and is widely used internationally due to it's robustness and accuracy \cite{SPANGFORS201662,Spangfors_NEWS,spagnolli}. NEWS evaluates patients according to their vital parameters, using values that are routinely monitored and utilized to develop a score for detecting patients at increased risk of mortality, cardiac arrest, and unplanned admission to the intensive care unit (ICU) within 24 hours \cite{NEWS-report}. NEWS has been validated in several studies, indicating that it is one of the most robust and accurate scores to identify these patients 
While finding patients at risk of adverse events is indeed useful, the extent to which these events are avoidable or unexpected is not evaluated in the development of these scores. We know that in-hospital death, which statistically is the main driver of NEWS alarms, are not unexpected or avoidable in the majority of cases. In a Norwegian study they found that less than 5 percent of in hospital deaths were considered avoidable after review \cite{Rogne49}. As such, NEWS seems not tuned or aimed toward detecting such events. Various machine learning approaches show promising results \cite{Che2018, shukla2018interpolationprediction}, but focus on prognosis of hard endpoints like death still receives most attention in the literature. As such, there seems to be a lack of research on characterizing clinical time-series for patients following expected normal trajectories from which abnormal deviations can be detected. Thus, our aim with this project is to initially identify such \textit{phenotype developments}. In this paper, we introduce a method based on a transformer architecture to cluster time-series from latent space representations. Utilizing this method, we identified a set of distinct clusters based on vitals recordings and static demographic information. In future work, we will further improve the model performance and evaluate clinical characterization of the clusters through the assessment of patients with medical experts and with new feature rich data collected from NHT.

\section{Related work}
The idea to detect characteristics of various patient groups clinical development is certainly not a new field of research. In general, the topic is associated with the study of time-series \cite{Ruiz2021}. At this stage, we cannot limit our training of the model to a supervised regime, due to the limited availability of such labeled data. Thus, we will further limit our research to  unsupervised model training of multivariate time-series \cite{Time-series_clustering}. This topic has been explored in a clinical context by for instance \cite{Marlin_Unsupervised_Pattern_Discovery}, where they utilize a mixture of diagonal covariance Gaussians to create a model for comparing physiological time-series. Such time-series often exhibit missing measurements, irregular sampling intervals, and correlations between parameters, making them challenging to analyze. To address these issues, statistical approaches such as the Gaussian Process view introduced in \cite{Li}, presents a kernel-based framework for classification of sparse and irregularly sampled time-series. These kernels produce similarity measures between time-series utilized in clustering algorithms, an essential ingredient in unsupervised classification. Additionally, handling missing data or non-aligned data-sampling were addressed in the work by \cite{OYVINDMIKALSEN2021107896}, who proposed time-series clustering kernels based on informative "missingness" of data. This approach exploited patterns in the missing data, often reflective of specific clinical conditions, to improve clustering accuracy.

With the advent of deep learning, particularly Recurrent Neural Networks (RNNs) like the Gated Recurrent Unit (GRU) \cite{cho-etal-2014-learning}, supervised learning of time-series data became feasible. These networks captured temporal dependencies as latent encodings, facilitating both predictions and classification tasks.  Shukla and Reddy \cite{shukla2018interpolationprediction} integrated interpolation networks with multivariate time-series, enabling information sharing across dimensions. A prominent example applied a GRU network in conjunction with a model for informative "missingness" to predict patient outcomes using the MIMIC-III and Physionet databases \cite{Che2018}. However, these approaches primarily focus on supervised learning, requiring labeled time-series data. 

For unsupervised learning with deep neural networks, autoencoders emerged as a valuable tool. Autoencoders are self-supervised models that learn to reconstruct input data from latent encoded representations, making them suitable for scenarios with limited labeled data. Bianchi et al. \cite{BIANCHI2019106973} introduced an autoencoder framework designed to handle missing values without imputation by aligning pairwise similarities in latent space representations using kernel functions \cite{MIKALSEN2018569}. 

In recent years, attention-based transformer networks have demonstrated remarkable success in various applications, particularly in natural language processing. Ienco and Interdonato \cite{AttentiveGatedAutoencoder} combined a recurrent autoencoder with attention-based enhancements to better utilize information across entire time-series for improved similarity measures in clustering. However, it struggled to handle irregular sampling intervals. This was addressed by the authors of \cite{SelfSupervisedTransformer}, taking the concept of transformers even further.  They consider each sample of the multivariate time-series as triples \((t_i,f_i,v_i)\), indicating $t_i$ as sampling time, $f_i$ as feature number and $v_i$ as the value. This representation effectively removes the need for imputation, treating each measurement as a unique data-point. The model supports a self-supervised objective, forecasting on training data for comparison with known future values. Additionally, the network employs fusion to concatenate demographic information like age, sex and comordabilities, typically being very important information for characterizing patient behavior. The network is coined Self-supervised Transformer for Time-Series (STraTS) and the paper is accompanied by published code available at \href{https://github.com/sindhura97/STraTS}{https://github.com/sindhura97/STraTS.
}
In summary, the field of adverse event detection in clinical time series has witnessed a significant evolution, from traditional methods to the transformative power of deep learning and attention-based transformers. These advancements have paved the way for more accurate and comprehensive analysis of patient data, offering new possibilities for improving healthcare outcomes.

\section{Method}
The main purpose of this paper is to develop and test methods for self-supervised learning of patient characteristic clinical developments. The motivation behind the architecture choice is reflected on the recent success of transformers on various applications for prediction. Moreover, there seem to be a lack of research on research of clustering medical time-series based on transformer-based encoders. In this regard, STraTS \cite{SelfSupervisedTransformer} is a promising candidate due to its success in predicting clinical outcomes based on complex multivariate time-series. As such, the objective was to modify STraTS to suit clustering applications and further test its performance for the clustering of the NHT-NEWS score time-series and compare it with a baseline method.

\subsection{The NEWS-NHT time-series}
Anonymous patient recordings for Vitala was extracted from the MEDANETS\textsuperscript{\textcopyright} system, encompassing data collected between January 2022 and April 2023. This dataset includes NEWS time-series recordings from 22,024 patients across multiple locations within the Nordland Hospital Trust (NHT). 
According to the NEWS system \cite{NEWS}, recordings for each patient should be sampled at least four times per day. In reality, the data is irregularly sampled and with high variability to fit the need for observation for possible critical care intervention. The lengths of the time-series is also highly variant  
since it is dependent on the discharge time and sample rate for the patient. The data was made anonymous with respect to patient ID, replacing it with a randomly generated number.

For our analyses, we chose to focus on the six continuous data types available in the NEWS recordings \cite{NEWS} which were systolic and diastolic blood pressure, O2 level in the blood, respiration rate, body temperature and pulse. Prior to training the network, these variables were normalized by subtracting the ideal value from the NEWS score system as seen in Table \ref{tab:normalization}. Furthermore, the data was then standardized with 
\[
\text{Standardized value} = \frac{{x - \text{mean(data)}}}{{\text{std(data)}}}
\]
to make sure all variables have the same range of values. 

\begin{table}[]
\begin{tabular}{ll}
\textbf{Variable}        & \textbf{Optimal Value} \\ \hline
Temperature              & 37            \\
O2 saturation            & 96            \\
Pulse                    & 70            \\
Systolic blood pressure  & 120           \\
Diastolic blood pressure & 80            \\
Respiratory frequency    & 16           
\end{tabular}
\caption{Values used for normalizing the continuous data.}
\label{tab:normalization}
\end{table}

\begin{figure}[t]
    \centering
    \includegraphics[width=1 \linewidth]{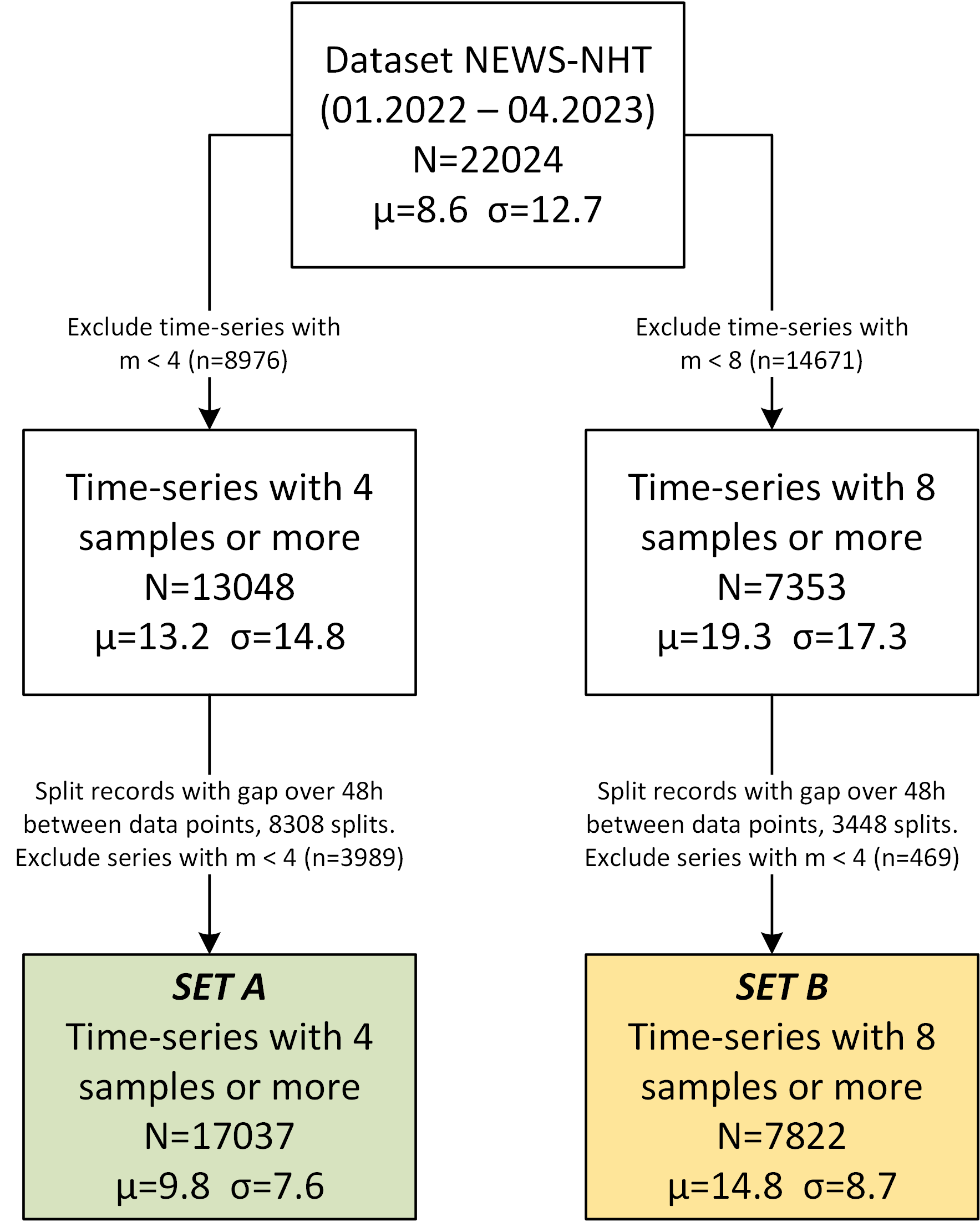}
    \caption{Workflow of filtering the dataset. Removing time-series with less than 4 and 8 data points for each patient. Also, we split time-series where patients have several visits (detected by time more than 48 hours between samples). N is number of series, $\mu$ and $\sigma$ relates to mean and standard deviation of lengths of series, $m$. }
    \label{fig:Data_scrubb}
\end{figure}

Patients with several separate visits with more than 48 hours in between were split and considered as separated patient time-series. To ensure sufficient number of data points for making any masked prediction, only patients with a minimum amount of data points were used. Two such minimum values were utilized, 4 and 8, to represent approximately 1 or 2 days of stay. 
This resulted in 2 datasets used in the experiments, an overview of how the data was divided can be seen in Figure \ref{fig:Data_scrubb}.

Around 24\% of the series had missing data for gender. The missing values were imputed with a random categorical value according to the distribution from the rest of the data set, with 51\% being males. 

\subsection{The modified STraTS Encoder}
\begin{figure*}[t]
    \centering
    \includegraphics[width=1 \linewidth]{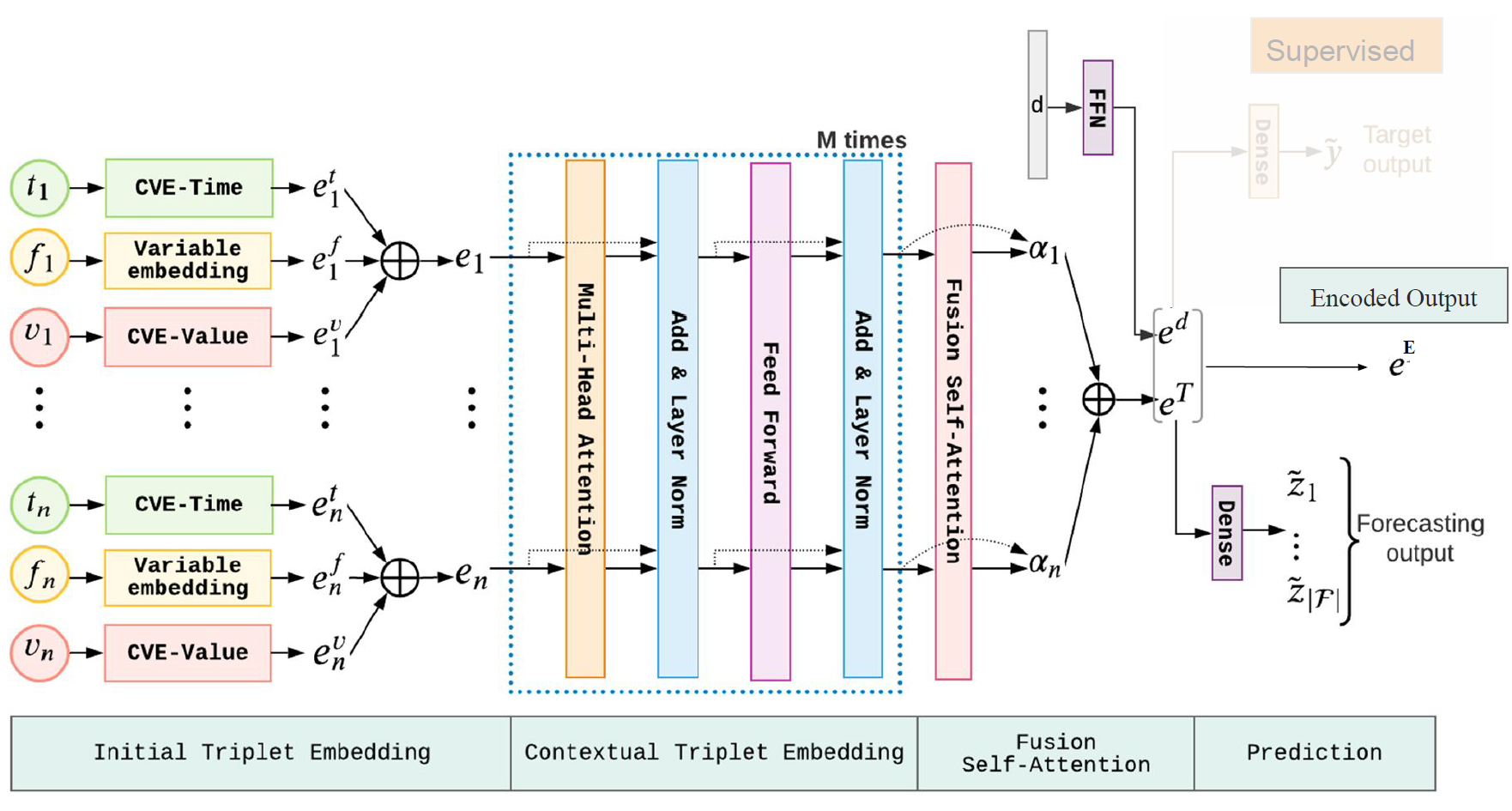}
    \caption{STraTS architecture with modifications to be used as an encoder by removing the supervised classification head (graphic reproduced with the kind permission from the authors of \cite{SelfSupervisedTransformer})}
    \label{fig:STraTS_acrhitecture}
\end{figure*}

STraTS is based on the encoder structure of the Transformer architecture \cite{SelfSupervisedTransformer}. This includes an initial embedding, a set of multi-head attention mechanisms, and an output prediction part, seen in Figure \ref{fig:STraTS_acrhitecture}. 

A certain time-series in NHT-NEWS can be described as $\mathbf{T}=\{(t_i,\mathbf{v}_i)\}_{i=1}^m$ , where $\mathbf{v}_i=[s_i,d_i, o_i, r_i, \tau_i, p_i, a_i]$ is the multivariate NEWS recording of vitalia consisting of $s_i$ being the systolic blood pressure, $d_i$ diastolic blood pressure, $o_i$ the oxygen saturation level in the blood, $r_i$ the respiration rate, $\tau_i$ the temperature in Celsius, $p_i$ the blood pressure and in addition $a_i$ being a categorical variable depicting whether extra oxygen is supplied for the patient. Furthermore, our dataset consists of $N$ such time-series with varying length $\mathcal{D}=\{\mathbf{T}^k\}_{k=1}^N$, where this length variation can be described as a set of $\{m_k\}_{k=1}^N$ for each time-series $\mathbf{T}^k=\{(t_i,\mathbf{v}_i)\}_{i=1}^{m_k}$. However, STraTS allows for independent time of sampling for individual variables, so each NEWS sample splits into 7 individual triplets. As such, we consider $n_k=7m_k$  such triplets $(v_i,f_i,t_i)$ for each each variable length time-series.

Furthermore, STraTS supports fusion of static demographic vectors for each patient time-series. In the NHT-NEWS dataset, information on gender and ward type were supplied as static data for each time-series. We refer to \cite{AttentiveGatedAutoencoder} for more details on the architecture of STraTS. The dimension of the encoded vector will be $d=d_{var} + d_{stat}$, where $d_{var}$ represents the dimensions of time-series embeddings, while $d_{stat}$ represents the dimensions of static demographic embeddings.

An \textit{Initial Triplet Embedding} is used to map each of the $n$ triplets of the input sequence into a vector $\mathbf{e_i} \in \mathbb{R}^{d_{var}}$, where $d_{var}$ is a hyper-parameter for the dimension of the mapped variables. The vector \(\mathbf{e_i}\) is calculated by summing all embeddings into a single embedding $\mathbf{e_i} = \mathbf{e_i^f} + \mathbf{e_i^v} + \mathbf{e_i^t}$. The embedding for the variable, $\mathbf{e_i^f}$, is found in a simple look-up table similar to word embeddings. For the value and time embeddings, $\mathbf{e_i^v}, \mathbf{e_i^t}$, we have continuous values. Here one to many Feed Forward Networks (FFN) with learnable parameters are used to calculate each embedding, $\mathbf{e_i^v} = FFN^v(v_i)$,  $\mathbf{e_i^t} = FFN^t(t_i)$. Both FFNs have a single input and a $d_{var}$ dimension output, there is one hidden layer with $\lfloor \sqrt{d_{var}} \rfloor$ nodes and a $tanh(\cdot)$ activation function. The embedding is then passed to the \textit{Contextual Tripled embedding} that consists of $M$ blocks with a multi-head attention (MHA) layer and a feed-forward layer, both followed by layer normalization. From the last MHA block, the output, $\mathbf{c_i}$ is then passed to the \textit{Fusion Self attention}. Here a FFN is used on every contextual embedding, $\mathbf{c_i}$, followed by a softmax to calculate attention weights, $\alpha_i$. The final embedding for the time-series can then be computed by a weighted sum of the contextual triplet embedding and the attention weights $e^T = \sum \alpha_i \mathbf{c_i}$.
For the demographic variables, the embedding, $\mathbf{e^d}$, is done with a FFN with 2 hidden layers with the size of $(2d_{stat}\times d_{stat})$ nodes, where $d_{stat}$ is the hyper-parameter controlling the size of the embedded demography vector.  As such, we encode variable length time-series to a vector of dimension $d$, where $d=d_{var}+d_{stat}$, are the sum of dimensions for the variable and demographic inputs.

To make a prediction from the encoded input, the demographic and time-series embeddings are concatenated to $\mathbf{e^E} = concat(\mathbf{e^T}, \mathbf{e^d})\in \mathbb{R}^d$ and passed through a dense layer with a Sigmoid activation function. 



\subsection{Modifications to the STraTS model}
For our problem, we are only interested in the unsupervised part of the network and therefore the second phase of the STraTS network is removed in our adaptation of the network. Instead, the encoded output is used for clustering so that trends in the data can be found. 
To do this the encoded output from the variable time-series data $e^T$ and the static demographic data $e^d$ are concatenated to $e^E$. This is used as input to a clustering method instead of a dense classification layer, the difference can be seen in Equation \ref{eq:changeSTraTS}
\begin{equation}
\begin{split}
    \text{original} &= \text{dense}(\text{concat}(e^T, e^d)) \\
    \text{ours} &= \text{clustering}(\text{concat}(e^T, e^d))
\end{split}
\label{eq:changeSTraTS}
\end{equation}
Since the NHT-NEWS dataset has limited amount of variables the input dimensions of the STraTS network are reduced for the variable time-series/tatic demographic data. From the original  50/50 to 40/10 input dimensions.

\subsection{Clustering}
To cluster the data various techniques were used, including K-means, Hierarchical Density-Based Spatial Clustering of Applications with Noise (HDBSCAN), Spectral Clustering (SC), and Gaussian Mixture Models (GMM). The results were analyzed using the Silhouette score and by visual inspection after using PCA as a dimension-reduction technique on the encoded data. To make a fair comparison of the Silhouette score both PCA and t-Distributed Stocastic Neighbor Embedding (t-SNE) reduced the dimensions to 3. For the STraTS encoder we also made experiments with PCA dimension reduction to 3 prior to clustering.

\section{Results}
In this section, we describe how we utilized the methodological framework described above to detect clusters from the NEWS-NHT dataset.  

\subsection{Baseline}
Initially, we established a baseline using PCA and T-SNE on a statistical descriptor of the data prior to various clustering techniques were employed. The minimum, maximum and mean values were considered for each patient time-series across every variable in the dataset. Subsequently, we applied clustering utilizing the mentioned techniques. Utilizing this method we did not find any clearly separable cluster structures, justifying a consideration of the temporal dimension through the modified STraTS encoder.


\begin{figure}[h]
    \centering
    \includegraphics[width=1\linewidth]{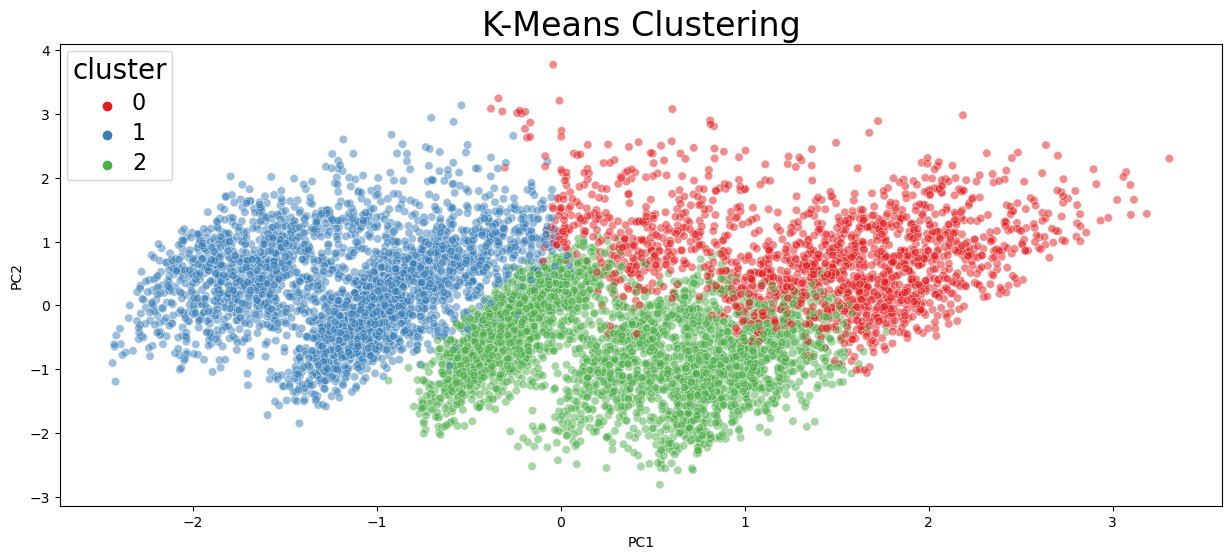}
    \caption{STraTS encoding with PCA before K-Means clustering on dataset B ($m \geq 8)$. Silhouette score 0.307.  }
    \label{fig:STraTS_PCA_KMeans}
\end{figure}


\begin{figure}[h]
    \centering
    \includegraphics[width=1.0\linewidth]{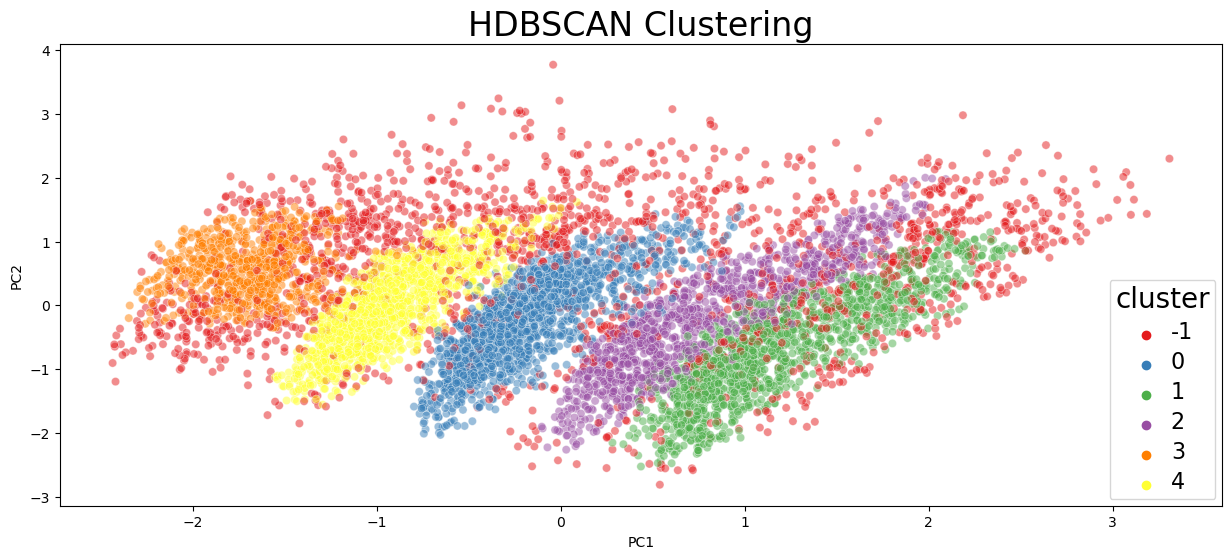}
    \caption{STraTS encoding with PCA before HDBSCAN clustering on dataset B ($m \geq 8$). Silhouette score 0.117.}
    \label{fig:STraTS_PCA_HDBSCAN}
\end{figure}


\begin{figure}[h]
    \centering
    \includegraphics[width=1.0\linewidth]{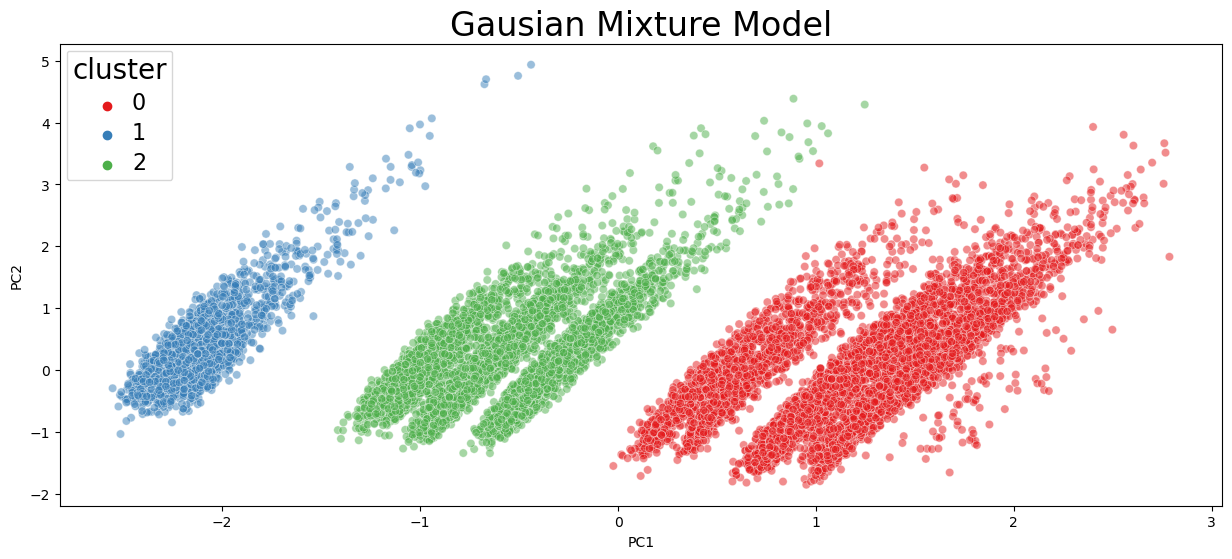}
    \caption{STraTS encoding with PCA before clustering using GMM with random init on dataset A  ($m \geq 4$), Silhouette score was 0.344 }
    \label{fig:STraTS_PCA_GMM}
\end{figure}


\begin{table*}[t]
\centering
\begin{tabular}{llllllllll}
\textbf{Model}    & \multicolumn{4}{l}{\textbf{Hyper parameters}}                                                                           &  \multicolumn{4}{l}{\textbf{Silhouette score Clustering Method}}  & \\ \hline
\textbf{Baseline} & \multicolumn{2}{l}{ }                                                  & Set           & Min, Max, Mean      & \textbf{Clusters}  & \textbf{KMeans} & \textbf{SC} & \textbf{GMM} &\textbf{HDB}    \\ \hline
PCA            &                                                         &                 & A             &   Max               & 3                  & 0.304           & 0.248       & 0.249        &  -- \\
T-SNE          &                                                         &                 & A             &   Max               & 3                  & 0.341           & 0.309       & 0.305        & -0.034\\
PCA            &                                                         &                 & B             &   Max               & 3                  & 0.303           & 0.266       & 0.250        &  -- \\
T-SNE          &                                                         &                 & B             &   Max               & 3                  & \textbf{0.353}           & 0.320       & 0.300        & 0.035  \\
               &                                                         &                 &               &                     &                    &                 &             &              &   \\
\textbf{\begin{tabular}[c]{@{}l@{}}STraTS\\ Encoder\end{tabular}}&\textbf{$d_{var}$}& \textbf{$d_{stat}$}& Set & \textbf{\begin{tabular}[c]{@{}l@{}}Demographic\\ variables\end{tabular}}& \textbf{Clusters}  & \textbf{KMeans} & \textbf{SC} & \textbf{GMM} &\textbf{HDB}\\ \hline
      & 40                                                                 & 10                  & A            & 1,2,3,4           & 3               & 0.232           & 0.212       & 0.197           & 0.113  \\
+ PCA & 40                                                                 & 10                  & A            & 1,2,3,4           & 3               & 0.348           & 0.348       & \textbf{0.349}           & 0.344  \\
      & 40                                                                 & 10                  & B            & 1,2,3,4           & 3               & 0.171           & 0.038       & 0.150           & -0.050 \\
+ PCA & 40                                                                 & 10                  & B            & 1,2,3,4           & 3               & 0.307           & 0.186       & 0.232           & 0.117 \\
\end{tabular}
\caption{Silhouette score for different clustering methods. T-SNE was made with 1000 iterations, 100 perplexity and with PCA initiation. Both PCA and T-SNE reduced the number of dimensions to 3. Demographic variables are 1 Gender, 2 Imputed Gender, 3 Ward type, 4 Change of ward during stay. No valid score is marked with $--$}
\label{tab:Silhouette score}
\end{table*}

\subsection{Temporal encoding}
The STraTS encoder \cite{SelfSupervisedTransformer}  without the classification head was used to encode the data before clustering it to a fixed-length vector with output dimensions $d$. Since the output is concatenated from the variable vitals measurements and the static demographic variables of dimension $d_{var}$ and $d_{stat}$, a range of different features and hyper-parameters were considered to better represent the complexity of the input configurations. 
During our experiments, we used a minimum length of each time-series of either 4 or 8 data points, representing 1 or 2 days with optimal sample rate (see Figure \ref{fig:Data_scrubb}). This ensured we had data both for masking and predicting in the training phase. We also removed the longest time-series with over 60 data points ($\approx1\%$ of patients) as they were considerably longer than the other time-series. We set the number of time-steps to predict to 1 as this should be the most straightforward task with limited data available. We also tried to predict more time steps without any improvement in clustering capabilities. 
As for the demographic vector, we tried different combinations of which variable types to include, in which "Gender" and "Ward type" was the most obvious. Some of the series lacked information on gender, making imputation necessary. Information on the occurrence of this was added in the binary variable "Imputed Gender", in line with the use of the "missingness" indicator mentioned in \cite{SelfSupervisedTransformer}. From a clinical perspective, we also introduced the binary variable "Change of ward during stay" describing if such an event was detected in the time-series.
To make a fair comparison between the different settings we left the following hyper-parameters constant for all experiments:  learning rate: 0.0005, batch size: 32, samples per epoch: 10240, patience: 5, MHA blocks in Transformer: 2, Heads in Transformer: 4, dropout: 0.2, max length of time-series: 60.

The resulting clustering scores for some hyper-parameter variations are presented in Table \ref{tab:Silhouette score}. Here we show the results from clustering baseline and STraTS embeddings. For the STraTS encoding we considered clustering on raw encoded vectors ($d=40$) in addition to cluster the three dominant PCA dimensions only, symbolized with +PCA in Table \ref{tab:Silhouette score}. 
We show illustrative results from the experiments in Figures \ref{fig:STraTS_PCA_KMeans}, \ref{fig:STraTS_PCA_HDBSCAN} and \ref{fig:STraTS_PCA_GMM}. Various clustering techniques were used on encoded STraTS + PCA embeddings, highlighted in figure caption. All experiments included fusion of the four mentioned variables in the demographic vector. Encoding dimensions was $d_{var}=40$ and $d_{stat}=10$. 

\section{Discussion and Conclusion}
\begin{figure}[hbt!]
    \centering
    \includegraphics[width=1.0\linewidth]{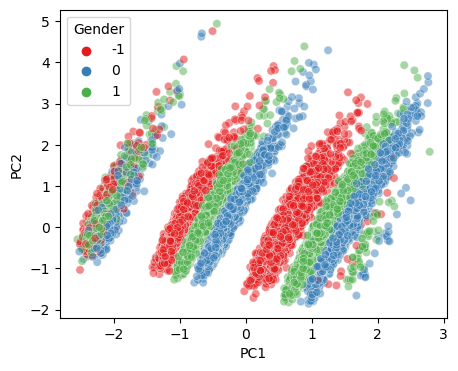}
    \caption{Gender colored overlay on encoded data}
    \label{fig:Gender_colormap}
\end{figure}

As our objective is the unsupervised learning of characteristic time-series development, we sought to identify hyper-parameter settings resulting in optimal clustering. As we see from the representative scatter plots in the Results section, there seem to be a certain degree of clustering tendencies. In Figure \ref{fig:STraTS_PCA_GMM} for $m \geq 4$, we can see clearly distinguishable clusters appear, successfully labeled by the GMM clustering algorithm. For the case of $m \geq 8$ depicted in figures \ref{fig:STraTS_PCA_KMeans} and \ref{fig:STraTS_PCA_HDBSCAN}, these clusters are far less distinct, but seem to have some of the same underlying topology.  For the $m \geq 8$ dataset there is only $N=7822$  series available, while for the $m \geq 4$, there is a far greater number $N=17037$, so one could speculate on statistical robustness for explanation. However, a more likely explanation is the contribution of the demographic information. When the number of samples in the time-series decreases, so does the influence of transformer part of network in favor of the demographic vector. This is due to less prominence of time-series prediction when sample-sizes are small. The phenomena can be witnessed in Figure \ref{fig:Gender_colormap}, where we have labeled the gender-information with a color code on each time-series. Since the genders are distinguishable as internal structures inside each cluster, we may also suspect the main clusters to reflect other such categorical variables in the demography vector. However, we have not as yet been able to identify this, so there is a certain probability for phenotypical behavior being reflected in these structures. This should be subject for further investigation both from a machine learning perspective, but also by a clinician which we at the time of writing have not been able to attain. In future work, we plan to collect a more feature rich dataset from the hospital, perform further training and model evaluation along with investigating the clinical characteristics of the clusters through patients assessments with medical experts. 

We conclude that these experiments has shown promising results emerging from a novel transformer-based model, but we believe that a more diverse data material should be considered, requiring more in-depth demographic knowledge about the patients like age and comorbidities. This aligns with the results in \cite{li2020application} where a Seq2Seq autoencoder was employed on similar data, having access to 75.762 time-series. The results indicates a lack of direct phenotypical information when only vitalia measurements are considered. Information from patients journal may supply the model with information about patient characteristics, currently not accessible for machine learning processing due to privacy constraints. This may include data on suspected diagnosis, treatment initiated at admission, medication and number of earlier admissions. Lastly, a more in-depth supervision of a dedicated clinician or medical researcher should be targeted in future studies. This may aid in the development of semi-supervised models, suitable for embedding expert advice on typical clinical developments in patient groups. Considering the richness and complexity of possible human clinical conditions, it seems natural to take into account previous medical knowledge in future iterations of developing anomaly detection systems based on clinical time-series.

\bibliographystyle{abbrvnat}
\Urlmuskip=0mu plus 1mu\relax
\bibliography{references}

\end{document}